\documentclass[letterpaper, 10 pt, conference]{ieeeconf}  

\IEEEoverridecommandlockouts                              
\usepackage[utf8]{inputenc}
\usepackage[T1]{fontenc}
\usepackage{multirow}
\usepackage{algorithm,algorithmic}
\usepackage{amssymb}
\usepackage[fleqn]{amsmath}
\usepackage{url}
\usepackage{graphicx}
\usepackage{etoolbox}
\makeatletter
\patchcmd{\@makecaption}
  {\scshape}
  {}
  {}
  {}
\makeatletter
\patchcmd{\@makecaption}
  {\\}
  {.\ }
  {}
  {}
\makeatother

\title{\LARGE \bf
Simple iterative method for generating targeted universal adversarial perturbations
}

\author{Hokuto Hirano$^{1}$ and Kazuhiro Takemoto$^{1,2}$
\thanks{$^{1}$Department of Bioscience and Bioinformatics, Kyushu Institute of Technology, Iizuka, Fukuoka 820-8502, Japan}%
\thanks{$^{2}$Corresponding author
        {\tt\small takemoto@bio.kyutech.ac.jp}}%
}

\begin{document}

\maketitle
\thispagestyle{empty}
\pagestyle{empty}

\begin{abstract}

Deep neural networks (DNNs) are vulnerable to adversarial attacks.
In particular, a single perturbation known as the universal adversarial perturbation (UAP) can foil most classification tasks conducted by DNNs.
Thus, different methods for generating UAPs are required to fully evaluate the vulnerability of DNNs.
A realistic evaluation would be with cases that consider targeted attacks; wherein the generated UAP causes DNN to classify an input into a specific class.
However, the development of UAPs for targeted attacks has largely fallen behind that of UAPs for non-targeted attacks. Therefore, we propose a simple iterative method to generate UAPs for targeted attacks. Our method combines the simple iterative method for generating non-targeted UAPs and the fast gradient sign method for generating a targeted adversarial perturbation for an input. We applied the proposed method to state-of-the-art DNN models for image classification and proved the existence of almost imperceptible UAPs for targeted attacks; further, we demonstrated that such UAPs are easily generatable.

\end{abstract}

\section{Introduction}
Deep neural networks (DNNs) are widely used for image classification, a task in which an input image is assigned a class from a fixed set of classes. For example, DNN-based image classification has applications in medical science (e.g., medical image-based diagnosis [1]) and self-driving technology (e.g., detecting and classifying traffic signs [2]).

However, DNNs are known to be vulnerable to adversarial examples [3], which are input images that cause misclassifications by DNNs and are generally generated by adding specific, imperceptible perturbations to original input images that have been correctly classified using DNNs. Interestingly, a single perturbation that can induce DNN failure in most image classification tasks is also generatable as a universal adversarial perturbation (UAP) [4]. The vulnerability in DNNs to adversarial attacks (UAPs, in particular) is a security concern for practical applications of DNNs [5]. Thus, the development of methods for generating UAPs is required to evaluate the vulnerability of DNNs to adversarial attacks.

A simple iterative method [4] for generating UAPs has been proposed; however, it is limited to non-targeted attacks that cause misclassification (i.e., a task failure resulting in an input image being assigned an incorrect class). More realistic cases need to consider targeted attacks, wherein generating a UAP would cause the DNN to classify an input image into a specific class (e.g., into the “diseased” class in medical diagnosis). A method for generating UAPs for targeted attacks based on a generative network model has been proposed [6]; however, it requires high computational costs. The targeted adversarial patch approach for targeted universal adversarial attacks [7] has been proposed; however, such adversarial patches are perceptible.

Thus, herein, we propose a simple iterative method to generate almost imperceptible UAPs for targeted attacks.

\section{Targeted universal adversarial perturbations}

Our algorithm (Algorithm 1) for generating UAPs for targeted attacks is an extension of the simple iterative algorithm for generating UAPs for non-targeted attacks [4]. Similar to the non-targeted UAP algorithm, our algorithm considers a classifier $C(\boldsymbol{x})$ that returns the class or label (with the highest confidence score) for an input image $\boldsymbol{x}$. The algorithm starts with $\boldsymbol{\rho}=\boldsymbol{0}$ (no perturbation) and iteratively updates the UAP $\boldsymbol{\rho}$ under the constraint that the $L_p$ norm of the perturbation is equal to or less than a small value $\xi$ (i.e., $\|\boldsymbol{\rho}\|_p\leq\xi$) by additively obtaining an adversarial perturbation for an input image $\boldsymbol{x}$, which is randomly selected from an input image set $\boldsymbol{X}$ without replacement. These iterative updates continue up till the termination conditions have been satisfied. Unlike the non-targeted UAP algorithm, our algorithm uses the fast gradient sign method for targeted attacks (tFGSM) to generate targeted UAPs, whereas the non-targeted UAP algorithm uses a method (e.g., DeepFool [8]) that generates a non-targeted adversarial example for an input image.

\begin{algorithm}[htbp]
\caption{Computation of a targeted UAP}
\begin{algorithmic}[1]
 \renewcommand{\algorithmicrequire}{\textbf{Input:}}
 \renewcommand{\algorithmicensure}{\textbf{Output:}}
 \REQUIRE Set $\boldsymbol{X}$ of input images, target class $y$, classifier $C(\cdot)$, cap $\xi$ on $L_p$ norm of the perturbation, norm type $p$ (1, 2, or $\infty$), maximum number $i_{\max}$ of iterations.
 \ENSURE  Targeted UAP vector $\boldsymbol{\rho}$.
  \STATE $\boldsymbol{\rho} \leftarrow \boldsymbol{0}$, $r_{st} \leftarrow 0$, $i \leftarrow 0$ 
  \WHILE {$r_{st}<1$ and $i<i_{\max}$}
  \FOR {$\boldsymbol{x}\in \boldsymbol{X}$ in random order}
  \IF {$C(\boldsymbol{x} + \boldsymbol{\rho}) \neq y$}
  \STATE $\boldsymbol{x}_{\mathrm{adv}}\leftarrow \boldsymbol{x}+\boldsymbol{\rho} + \psi(\boldsymbol{x} + \boldsymbol{\rho},y)$
  \IF {$C(\boldsymbol{x}_{\mathrm{adv}})=y$}
  \STATE $\boldsymbol{\rho} \leftarrow \mathrm{project}(\boldsymbol{x}_{\mathrm{adv}}-\boldsymbol{x},p,\xi)$
  \ENDIF
  \ENDIF
  \ENDFOR
  \STATE $r_{st} \leftarrow |\boldsymbol{X}|^{-1}\sum_{\boldsymbol{x}\in\boldsymbol{X}}\mathbb{I}\left(C(\boldsymbol{x}+\boldsymbol{\rho})=y\right)$
  \STATE $i\leftarrow i + 1$
  \ENDWHILE
 \end{algorithmic} 
\end{algorithm}

tFGSM generates a targeted adversarial perturbation $\psi(\boldsymbol{x},y)$ that causes an image $\boldsymbol{x}$ to be classified into the target class $y$ using the gradient $\nabla_{\boldsymbol{x}}{\cal L}(\boldsymbol{x},y)$ of the loss function with respect to pixels [3,9].
For the $L_{\infty}$ norm, the perturbation is calculated as
\begin{equation}
\psi(\boldsymbol{x},y) = -\epsilon \cdot \mathrm{sign}\left(\nabla_{\boldsymbol{x}}{\cal L}(\boldsymbol{x},y)\right),
\end{equation}
where $\epsilon$ ($>0$) the attack strength.
For the $L_1$ and $L_2$ norms, the perturbation is obtained as
\begin{equation}
\psi(\boldsymbol{x},y) = -\epsilon \frac{\nabla_{\boldsymbol{x}}{\cal L}(\boldsymbol{x},y)}{\|\nabla_{\boldsymbol{x}}{\cal L}(\boldsymbol{x},y)\|_p}.
\end{equation}
The adversarial example $\boldsymbol{x}_{\mathrm{adv}}$ is obtained as follows:
\begin{equation}
\boldsymbol{x}_{\mathrm{adv}} = \boldsymbol{x} + \psi(\boldsymbol{x},y).
\end{equation}

At each iteration step, our algorithm computes a targeted adversarial perturbation $\psi(\boldsymbol{x}+\boldsymbol{\rho},y)$, if the perturbated image $\boldsymbol{x}+\boldsymbol{\rho}$ is not classified into the target class $y$ (i.e., $C(\boldsymbol{x}+\boldsymbol{\rho})\neq y$); however, the non-targeted UAP algorithm obtains a non-targeted adversarial perturbation that satisfies $C(\boldsymbol{x}+\boldsymbol{\rho})\neq C(\boldsymbol{x})$ if $C(\boldsymbol{x}+\boldsymbol{\rho})=C(\boldsymbol{x})$.
After generating the adversarial example at this step (i.e., $\boldsymbol{x}_{\mathrm{adv}}\leftarrow \boldsymbol{x}+\boldsymbol{\rho} + \psi(\boldsymbol{x} + \boldsymbol{\rho},y))$, the perturbation $\boldsymbol{\rho}$ is updated if $\boldsymbol{x}_{\mathrm{adv}}$ is classified into the target class $y$ (i.e., $C(\boldsymbol{x}_{\mathrm{adv}})=y$), whereas the non-targeted UAP algorithm updates the perturbation $\boldsymbol{\rho}$ if $C(\boldsymbol{x} + \boldsymbol{\rho}) \neq C(\boldsymbol{x})$.
Note that tFGSM does not ensure that adversarial examples are classified into a target class. When updating $\boldsymbol{\rho}$, a projection function $\mathrm{project}(\boldsymbol{x},p,\xi)$ is used to satisfy the constraint that $\|\boldsymbol{\rho}\|_p \leq \xi$ (i.e., $\boldsymbol{\rho} \leftarrow \mathrm{project}(\boldsymbol{x}_{\mathrm{adv}}-\boldsymbol{x},p,\xi)$.
This projection is defined as follows:
\begin{equation}
\mathrm{project}(\boldsymbol{x},p,\xi) = \arg \underset{\boldsymbol{x'}}{\min} \| \boldsymbol{x} - \boldsymbol{x'}\|_2 \ \ \mathrm{s.t.} \ \ \|\boldsymbol{x'}\|_p \leq \xi
\end{equation}

This update procedure terminates when the targeted attack success rate $r_{ts}$ for input images (i.e., the proportion of input images classified into the target class; $|\boldsymbol{X}|^{-1}\sum_{\boldsymbol{x}\in\boldsymbol{X}}\mathbb{I}\left(C(\boldsymbol{x}+\boldsymbol{\rho})=y\right)$ equals 100\% (i.e., all input images are classified into the target class due to the UAP $\boldsymbol{\rho}$) or the number of iterations reaches to the maximum $i_{\max}$.

A pseudo code of our algorithm is shown in Algorithm 1.

Our algorithm was implemented using Keras (version 2.2.4; keras.io) and Adversarial Robustness 360 Toolbox [9] (version 1.0; \url{github.com/IBM/adversarial-robustness-toolbox}). The source code of our proposed method for generating targeted UPAs is available from our GitHub repository: \url{github.com/hkthirano/targeted_UAP_CIFAR10}.

\section{Experimental evaluation}
\subsection{Deep neural network models and image datasets}
To evaluate targeted UAPs, we used 2 DNN models that were trained to classify the CIFAR-10 image dataset (\url{www.cs.toronto.edu/~kriz/cifar.html}). The CIFAR-10 dataset includes 60,000 RGB color images with size of $32\times 32$ pixels classified into 10 classes: airplane, automobile, bird, cat, deer, dog, frog, horse, ship, and truck. 60,000 images are available in each class. The dataset comprises 50,000 training images (5,000 images per class) and 10,000 test images (1,000 images per class). In particular, we used the VGG-20 and ResNet-20 models for the CIFAR-10 dataset obtained from a GitHub repository (\url{github.com/GuanqiaoDing/CNN-CIFAR10}); their test accuracies were 91.1\% and 91.3\%, respectively.

Moreover, we also considered three DNN models trained to classify the ImageNet image dataset (\url{www.image-net.org}). The ImageNet dataset comprises RGB color images with size of $224 \times 224$ pixels classified into 1,000 classes. In particular, we used the VGG-16, VGG-19, and ResNet-50 models for ImageNet dataset available in Keras (version 2.2.4; \url{keras.io}), and their test accuracies were 71.6\%, 71.5\%, and 74.6\%, respectively.

\subsection{Generating targeted adversarial perturbations and evaluating their performance}

Targeted UAPs were generated using an input image set obtained from the datasets. The parameters p was set to 2. We generated targeted UAPs with various norms by adjusting the parameters $\epsilon$ and $\xi$. The magnitude of a UAP was measured using a normalized $L_2$ norm of the perturbation; in particular, we used the ratio $\zeta$ of the $L_2$ norm of the UAP to the average $L_2$ norm of an image in a dataset. The average $L_2$ norms of an image were 7,381 and 50,135 in the CIFAR-10 and ImageNet datasets, respectively.

For comparing the performance of targeted UAPs generated by our method with random controls, we also generated random vectors (random UAPs) sampled uniformly from the sphere of a given radius [4].

The performance of UAPs was evaluated using the targeted attack success rate $r_{ts}$. In particular, we considered the success rates $r_{ts}$ for input images. In addition to this, we also computed the success rates $r_{ts}$ for test images to experimentally evaluate the performance of UAPs for unknown images. A test image set was obtained from the dataset and was not overlapped with the input image set.

\subsection{Case of CIFAR-10 models}
For the CIFAR-10 models, we used 10,000 input images to generate the targeted UAPs. The input image set was obtained by randomly selecting 1,000 images per class from the training images of the CIFAR-10 dataset. All 10,000 test images of the dataset were used as test images for evaluating the UAP performance. We considered the targeted attack to each class. The parameters $\epsilon$ and $i_{\max}$ were set to 0.006 and 10, respectively.

For the targeted attacks to each class, the targeted attack success rates $r_{ts}$ for both the input image set and the UAP test image set, rapidly increased with perturbation rate, despite a low  $\zeta$ (2--6\%). In particular, the success rates were $>80\%$ for $\zeta=5\%$ (Fig. 1).
The targeted UAPs with $\zeta=5\%$ were almost imperceptible (Fig. 2). Moreover, the UAPs seem to represent object shapes of each target class.
The target attack success rates reached to $\sim 100\%$ for $\zeta>10\%$. The success rates of the targeted UAPs were significantly higher than those of random UAPs. These tendencies were observed both in the VGG-20 model and in the ResNet-20 model.

\begin{figure}[htbp]
\begin{center}
\includegraphics[width=82mm]{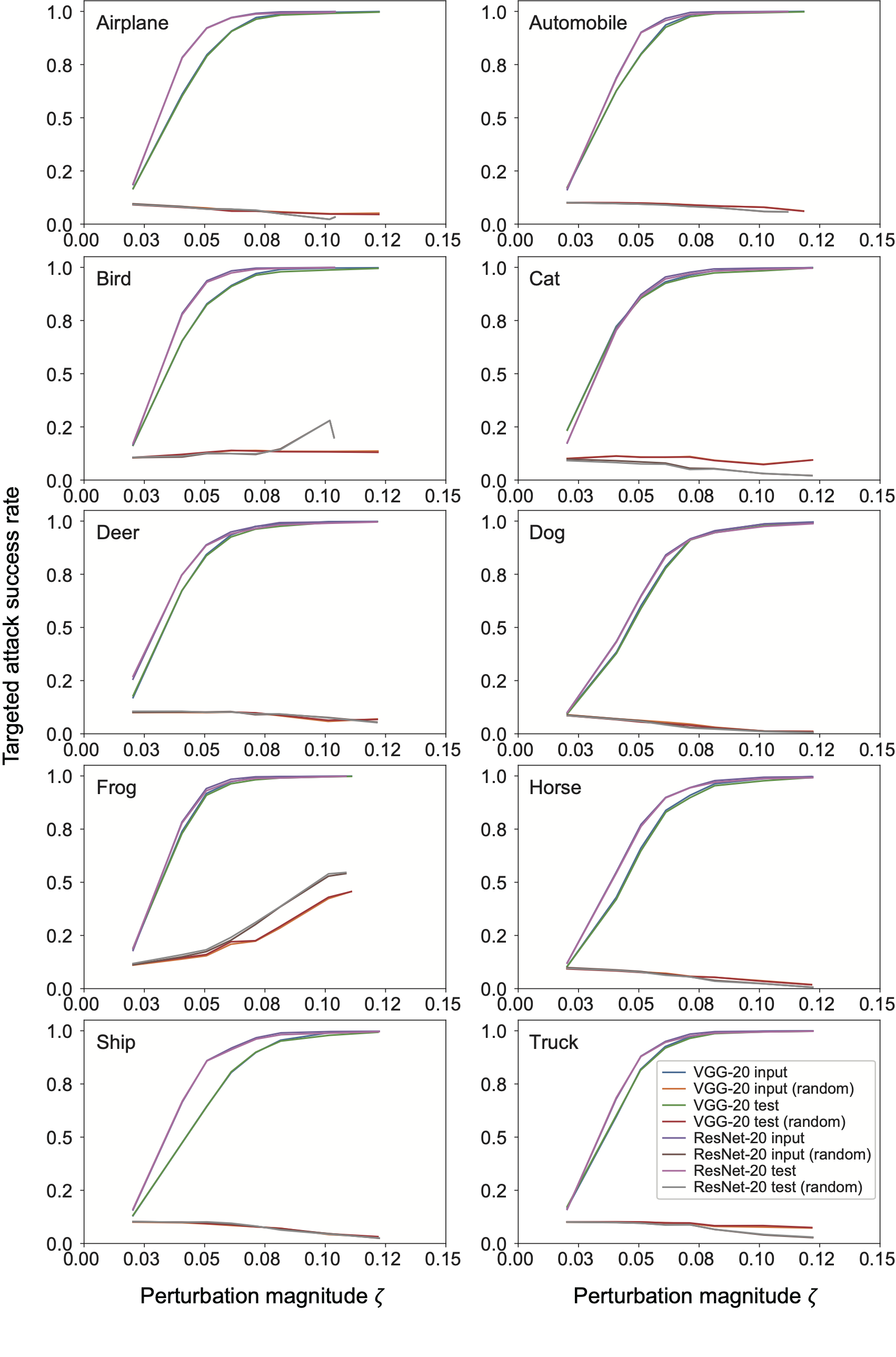} 
\caption{
Line plot of target attack success rate $r_{ts}$ versus perturbation rate for targeted attacks to each class of the CIFAR-10 dataset. Legend label indicates DNN model and image set used for computing $r_{ts}$. For example, ``VGG-20 input'' indicates $r_{ts}$ of targeted UAPs against the VGG-20 model computed using the input image set. Additional argument ``(random)'' indicates that random UAPs were used instead of targeted UAPs.
}
\end{center}
\end{figure}

\begin{figure}[htbp]
\begin{center}
\includegraphics[width=82mm]{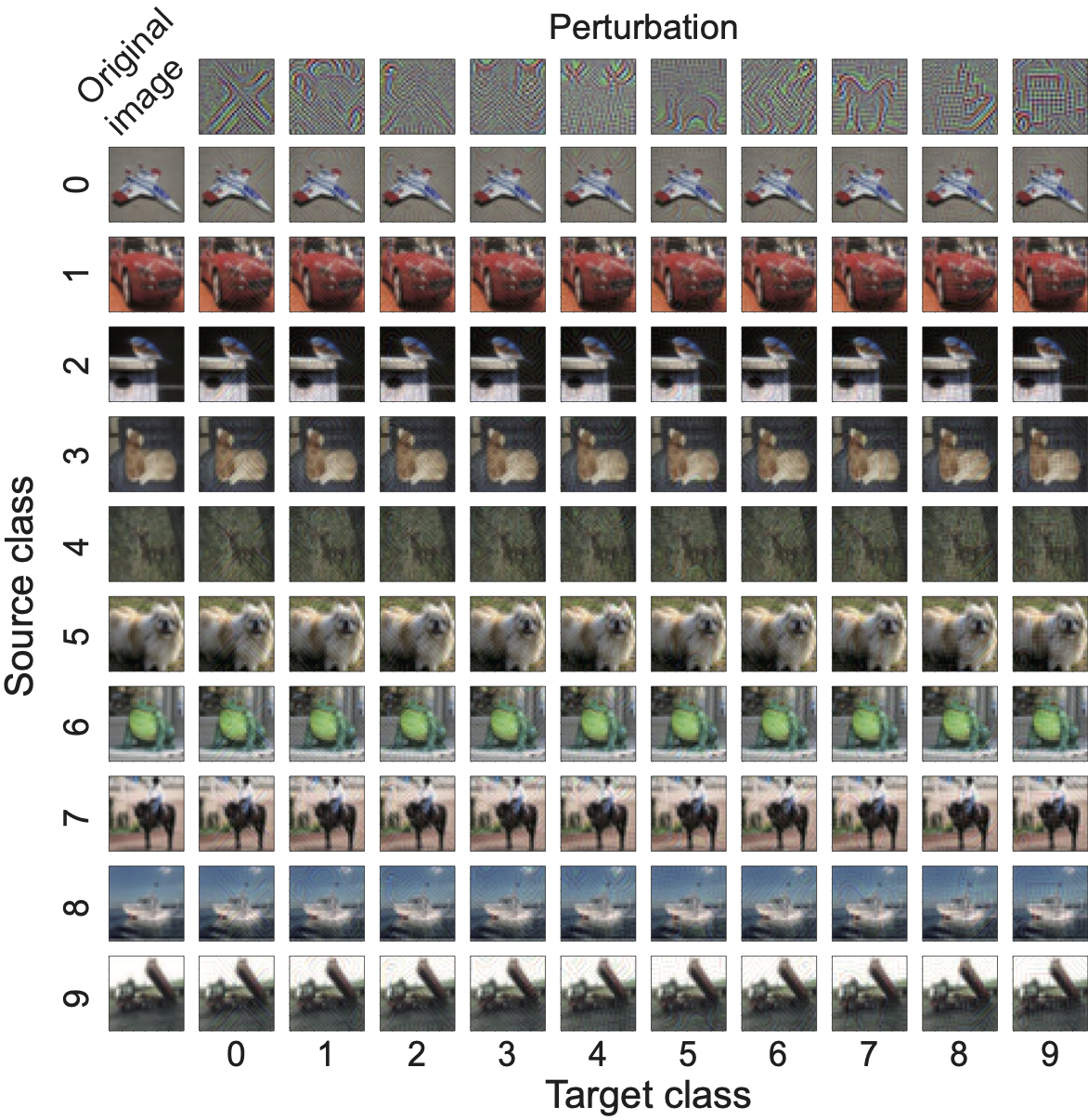} 
\caption{
Targeted UAPs (top panel) with $\zeta=5\%$ against the VGG-20 model for the CIFAR-10 dataset and their adversarial attacks to an original (i.e., non-perturbated) image (left panel) randomly selected from the images that, without perturbation, correctly classified into each source class and, with the perturbations, correctly classified into the target classes: airplane (0), automobile (1), bird (2), cat (3), deer (4), dog (5), frog (6), horse (7), ship (8), and truck (9). Note that the UAPs are emphatically displayed for clarity; in particular, each UAP was scaled with the maximum of 1 and the minimum of 0.
}
\end{center}
\end{figure}

\subsection{Case of ImageNet models}
For the ImageNet models, we used the validation dataset used in the ImageNet Large Scale Visual Recognition Challenge 2012 (ILSVRC2012; \url{www.image-net.org/challenges/LSVRC/2012/}) to generate the targeted UAPs. The dataset comprises 50,000 images (50 images per class). We used 40,000 images as input images. The input image set was obtained by randomly selecting 40 images per class. The rest (10,000 images; 10 images per class) was used as test images for evaluating UAPs. The parameters $\epsilon$ and $i_{\max}$ were set to 0.5 and 5, respectively.

In this study, we considered targeted attacks to three classes (golf ball, broccoli, and stone wall) that were randomly selected from 1,000 classes in a previous study [5] because of page limitation.

We generated targeted UAPs with $\zeta=6\%$ ($\xi=3,000$) and $\zeta=8\%$ ($\xi=4,000$). The target attack success rates $r_{ts}$ were between \textasciitilde30\% and \textasciitilde75\% and between \textasciitilde60\% and \textasciitilde90\% when $\zeta=6\%$ and $\zeta=8\%$, respectively (Table 1). The success rates of the targeted UAPs were significantly higher than those of random UAPs, which were less than 1\% in all cases.

\begin{table}[htbp]
\caption{Targeted attack success rates $r_{ts}$ of targeted UAPs against the DNN models for each target class. $r_{ts}$ for input images and test images were shown.}
\begin{center}
  \begin{tabular}{c|c|cc|cc}
  \hline
    \multirow{2}{*}{Target class} & \multirow{2}{*}{Model} & \multicolumn{2}{c|}{$\zeta=6\%$} & \multicolumn{2}{c}{$\zeta=8\%$}  \\
     & & input & test & input & test \\
     \hline
    \multirow{3}{*}{Golf ball} & VGG-16 & 58.0\% & 57.6\% & 81.6\% & 80.6\% \\
    & VGG-19 & 55.3\% & 55.2\% & 81.3\% & 80.1\% \\
    & ResNet-50 & 66.8\% & 66.5\% & 90.3\% & 89.8\% \\
    \hline
    \multirow{3}{*}{Broccoli} & VGG-16 & 29.3\% & 29.0\% & 59.7\% & 59.5\% \\
    & VGG-19 & 31.2\% & 30.5\% & 59.7\% & 59.4\% \\
    & ResNet-50 & 46.4\% & 46.6\% & 74.6\% & 73.9\% \\
    \hline
    \multirow{3}{*}{Stone wall} & VGG-16 & 47.1\% & 46.7\% & 75.0\% & 74.5\% \\
    & VGG-19 & 48.4\% & 48.1\% & 73.9\% & 72.9\% \\
    & ResNet-50 & 74.7\% & 74.4\% & 92.0\% & 91.3\% \\
    \hline
  \end{tabular}
  \end{center}
\end{table}

A higher perturbation magnitude $\zeta$ leaded to a higher targeted attack success rate $r_{ts}$. The success rates $r_{ts}$ depended on the image classes. For example, the targeted attacks to the class ``Golf ball'' were more easily achieved than those to the class ``Broccoli''. The success rates $r_{ts}$ also depended DNN architectures; in particular, the ResNet-50 model was more easy-to-fool than the VGG models.

The targeted UAPs with $\zeta=6\%$ and $\zeta=8\%$ were almost imperceptible (Fig. 3); however, they were partly perceptible for whitish images (e.g., trimaran). Moreover, the UAPs seem to reflect object shapes of each target class.

The targeted attack success rates in the ImageNet models were relatively lower than those in the CIFAR-10 models. This is because the ImageNet dataset has a larger number of classes than the CIFAR-10 dataset does. In short, it is more difficult to exactly classify an input image into a specific target class within a larger number of classes. Moreover, the observed lower success rate may be because the validation dataset of ILSVRC2012 was used when generating targeted UAPs. Higher success rates may be obtained when generating targeted UAPs using training images.

\begin{figure}[htbp]
\begin{center}
\includegraphics[width=82mm]{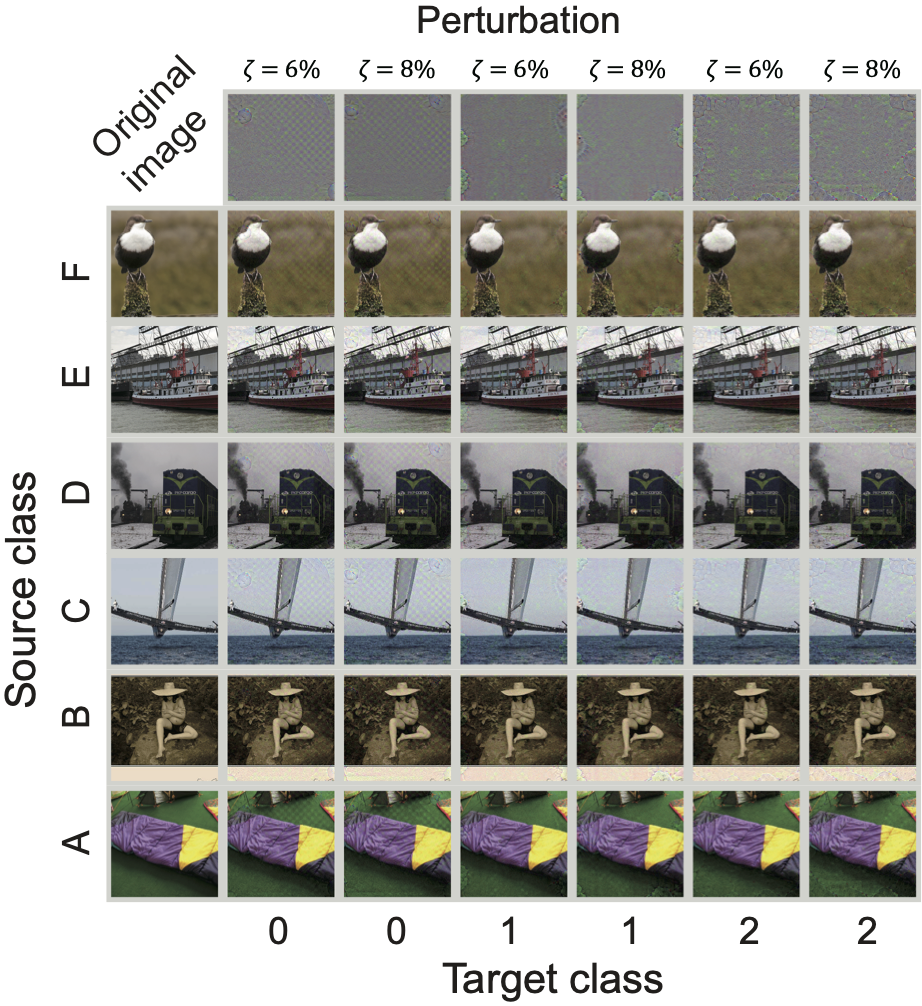} 
\caption{
Targeted UAPs (top panel) against the ResNet-50 model for the ImageNet dataset and their adversarial attacks to original (i.e., non-perturbated) images (left panel) randomly selected from the images that, without perturbation, correctly classified into the source class and, with the perturbation, correctly classified into each target classes under the constraint that the source classes are not overlapped each other and with the target classes. The source classes displayed here are sleeping bag (A), sombrero (B), trimaran (C), steam locomotive (D), fireboat (E), and water ouzel, dipper (F). The target classes are golf ball (0), broccoli (1), and stone wall (2). The UAPs with $\zeta=6\%$ and $\zeta=8\%$ are shown. Note that UAPs are emphatically displayed for clarity; in particular, each UAP was scaled with the maximum of 1 and the minimum of 0.
}
\end{center}
\end{figure}

\section{Conclusions}
We propose a simple iterative method to generate targeted UAPs for image classification, although the proposed algorithm is a straightforward extension of the non-targeted UAP algorithm [4]. A similar iterative algorithm has been proposed to generate targeted UAPs for audio processing systems [10].
Using the CIFAR-10 and ImageNet models, we demonstrated that a small (almost imperceptible) UAP generated by our method made the models largely classify test images into a target class. Our method is expected to have lower computational costs, compared with the generative model-based method [5]; however, such a comparison is a subject of future investigation.
Our results indicated the existence of UAPs for targeted attacks and that such UAPs are easily generatable. Our study enhances our understanding of the vulnerabilities of DNNs to adversarial attacks; moreover, it may help increase the security of DNNs [11].


\begin{thebibliography}{99}

\bibitem{}
Esteva A, et al (2017), Dermatologist-level classification of skin cancer with deep neural networks. Nature 542:115–118

\bibitem{}
Stallkamp J, et al (2013), Man vs. computer: Benchmarking machine learning algorithms for traffic sign recognition. Neural Networks 32:323--332

\bibitem{}
Goodfellow IJ, Shlens J, Szegedy C (2014), Explaining and harnessing adversarial examples. arXiv preprint arXiv:1412.6572

\bibitem{}
Moosavi-Dezfooli S-M, et al (2017), Universal adversarial perturbations. Proceedings of IEEE Conference on Computer Vision and Pattern Recognition (CVPR) 2017

\bibitem{}
Hayes J, Danezis G (2018) Learning universal adversarial perturbations with generative models. Proceedings of IEEE Symposium on Security and Privacy 2018

\bibitem{}
Finlayson SG, et al (2019) Adversarial attacks on medical machine learning. Science 363:1287–1289

\bibitem{}
Brown TB, et al (2017) Adversarial patch. arXiv preprint arXiv:1712.09665

\bibitem{}
Mohsen S, Fawzi A, Frossard P (2016) DeepFool: a simple and accurate method to fool deep neural networks. Proceedings of IEEE Conference on Computer Vision and Pattern Recognition (CVPR) 2016.

\bibitem{}
Nicolae M-I, et al (2018), Adversarial Robustness Toolbox v1.0.1. arXiv preprint arXiv:1807.01069

\bibitem{}
Abdoli S, et al (2019) Universal adversarial audio perturbations. arXiv preprint arXiv:1908.03173

\bibitem{}
Yuan X, et al (2019) Adversarial examples: attacks and defenses for deep learning. IEEE Transactions on Neural Networks and Learning Systems 30:2805--2824






\end{thebibliography}
\end{document}